%% file: main.tex
\documentclass{article}
\usepackage{spconf,amsmath,amssymb,graphicx,subcaption,multirow}

\def\x{{\mathbf x}}
\def\y{{\mathbf y}}
\def\R{{\mathbb{R}}}
\def\yhat{{\mathbf{\hat{y}}}}
\def\henc{{\mathbf h}^\text{enc}}

\def\blank{\left<\text{b}\right>}
\def\sos{\left<\text{sos}\right>}

\title{Less is More: Improved RNN-T Decoding using Limited Label Context and Path Merging}
\name{Rohit Prabhavalkar, Yanzhang He, David Rybach, Sean Campbell}
\secondlinename{Arun Narayanan, Trevor Strohman, Tara N. Sainath}
\address{Google, USA}
\begin{document}
\ninept
\maketitle
\begin{abstract}
	\input{abstract}
\end{abstract}
\input{introduction}
\input{rnnt}
\input{experiments}

\input{results}
\input{conclusions}

\vfill\pagebreak
\bibliographystyle{IEEEbib}
\bibliography{main}

\end{document}

%% file: abstract.tex
End-to-end models that condition the output label sequence on all previously
predicted labels have emerged as popular alternatives to conventional systems
for automatic speech recognition (ASR).
Since unique label histories correspond to distinct models states, such models
are decoded using an approximate beam-search process which produces a tree of
hypotheses.

In this work, we study the influence of the amount of label context on the
model's accuracy, and its impact on the efficiency of the decoding process.
We find that we can limit the context of the recurrent neural network transducer
(RNN-T) during training to just four previous word-piece labels, without
degrading word error rate (WER) relative to the full-context baseline.
Limiting context also provides opportunities to improve the efficiency of the
beam-search process during decoding by removing redundant paths from the active
beam, and instead retaining them in the final lattice.
This path-merging scheme can also be applied when decoding the baseline
full-context model through an approximation.
Overall, we find that the proposed path-merging scheme is extremely effective
allowing us to improve oracle WERs by up to 36\% over the baseline, while
simultaneously reducing the number of model evaluations by up to 5.3\% without
any degradation in WER.

%% file: introduction.tex
\section{Introduction}
\label{sec:intro}
End-to-end models such as the recurrent neural network transducer
(RNN-T)~\cite{Graves12, GravesMohamedHinton13, RaoSakPrabhavalkar17},
attention-based encoder-decoder models~\cite{ChanJaitlyLeEtAl16,
ChorowskiBahdanauSerdyukEtAl15}, the transformer
transducer~\cite{ZhangLuSakEtAl20, YehMahadeokarKalgaonkarEtAl19} have become
increasingly popular alternatives to conventional hybrid
systems~\cite{MorganBourlard95} for automatic speech recognition.
These models produce hypotheses in an autoregressive fashion by
conditioning the output on all previously predicted labels, thus making fewer
conditional independence assumptions than conventional hybrid systems.
End-to-end ASR models have been shown to achieve state-of-the-art
results~\cite{ParkChanZhangEtAl19, GulatiQinChiuEtAl20} on popular public
benchmarks, as well as on on large scale industrial
datasets~\cite{ChiuSainathWuEtAl18, SainathHeLiEtAl20}.

The increase in modeling power afforded by conditioning on all previous
predictions, however, comes at the cost of a more complicated decoding process;
computing the most likely label sequence \emph{exactly} is intractable since it
involves a discrete search over an exponential number of sequences each of which
corresponds to a distinct model state.
Instead, decoding is performed using an approximate beam
search~\cite{SutskeverVinyalsLe14}, with various heuristics to improve
performance~\cite{ChorowskiJaitly17, TripathiLuSakEtAl19}.
Since model states corresponding to different label histories are unique, beam
search decoding produces a \emph{tree of hypotheses} rooted at the start of
sentence label ($\sos$).
This, combined with a limited beam size, restricts the diversity of decoded
hypotheses -- a problem that becomes increasingly severe for longer utterances.

In this work, we conduct a detailed investigation of a specific aspect of
streaming end-to-end models (RNN-T, in our work); we study the importance of
conditioning the output sequence on the full history of previously predicted
labels, and investigate modifications to the beam search process which can be
applied to models by limiting context.
The first part of this question has been investigated in a few papers recently,
in different contexts.
Ghodsi \emph{et al.}~\cite{GhodsiLiuApfelEtAl20} find that in low-resource settings,
where training data is limited, the use of word-piece
units~\cite{SchusterNakajima12} allows for a stateless prediction network (i.e.,
one which conditions on only one previous label) without a significant loss
in accuracy.
Zhang \emph{et al.}~\cite{ZhangLuSakEtAl20} investigate the impact of varying label
context in the transformer-transducer model (RNN-T which replaces LSTMs with
transformer networks~\cite{VaswaniShazeerParmarEtAl17}) finding that a context
of 3-4 previous graphemes achieves similar performance as a full-context
baseline on the Librispeech dataset.
Finally, Variani \emph{et al.}~\cite{VarianiRybachAllauzenEtAl20} find that the
hybrid autoregressive transducer (HAT; RNN-T with an `internal language model
(LM)'), trained to output phonemes and decoded with a separate lexicon and
grammar achieves similar performance when context is limited to two previous
phoneme labels on a large scale task.\footnote{It should be noted, however, that
in this case the effective context is larger than two previous phonemes due to
the linguistic constraints introduced by the lexicon and the n-gram LM.}

Our work differs from the previously mentioned works in two ways.
First, we study the question on a large-scale task using a model trained to
output word-piece targets and decoded without an external lexicon or language
model.
Since word-pieces naturally capture longer context than graphemes of phonemes,
our results allow us to measure the effective context that is captured by
full-context RNN-T models.
Second, we consider modifications to the beam search process which are
enabled by the use of limited context models (or the baseline, with
approximations).
This process, described in detail in Section~\ref{sec:path_merging}, allows us to
generate lattices by merging paths during the search process, unlike RNN-T
systems which are typically decoded to produce trees.
The proposed approach is similar to previous work on efficient rescoring with
neural LMs~\cite{LiuWangChenEtAl14} and to generating lattices in
attention-based encoder-decoder models~\cite{ZapotocznyPietrzakLancuckiEtAl19}.
The ability to produce rich lattices from sequence-to-sequence models has many
potential applications: e.g., they can be used for spoken term
detection~\cite{MillerKleberKaoEtAl07}; as inputs to spoken language
understanding systems~\cite{DeMoriBechetHakkaniTurEtAl08};
or for computing word-level posteriors for word- or utterance-level confidence
estimation~\cite{KempSchaaf97}.
In experimental evaluations, we find that the models require 5-gram contexts
(i.e., conditioning on the four previously predicted labels) in order to obtain
comparable WER results as the baseline.
Additionally, we find that the proposed path-merging scheme is an effective
technique to improve search efficiency.
The proposed scheme reduces the number of model evaluations by up to 5.3\%, while
simultaneously improving the oracle WERs in the decoded lattice by up to
36\% without any degradation in the WER.

The organization of the rest of the paper is as follows: in
Section~\ref{sec:rnnt} we introduce the baseline RNN-T models; in
Section~\ref{sec:limited_context} we describe how we limit label context during
training, and our proposed path-merging scheme during decoding to produce
lattices; we describe our experimental setup and results in
Sections~\ref{sec:experiments} and~\ref{sec:results}, respectively, before
concluding in Section~\ref{sec:conclusions}.

%% file: rnnt.tex
\section{The Recurrent Neural Transducer (RNN-T)}
\label{sec:rnnt}
\begin{figure}
  \centering
  \includegraphics[width=0.85\columnwidth]{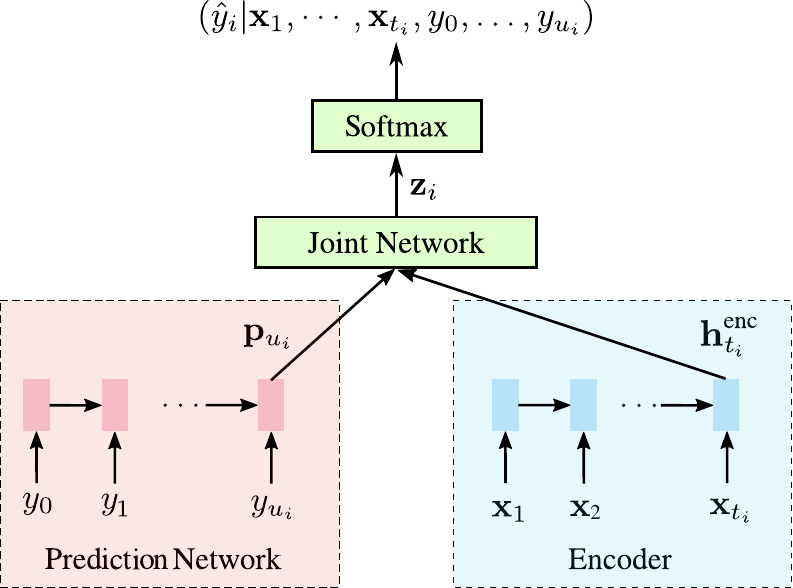}
  \caption{A representation of a (full-context) RNN-T model.}
  \label{fig:rnnt}
\end{figure}
We assume that the input speech utterance has been parameterized into suitable
acoustic features: $\x = [\x_1, \ldots, \x_T]$, where $\x_t \in \R^d$ (log-mel
filterbank energies, in this work).
Each utterance has a corresponding label sequence, $\y = [y_1, \ldots, y_U]$,
where $y_u \in \mathcal{Y}$ (word-pieces~\cite{SchusterNakajima12}, in this
work).

The RNN-T model was proposed by Graves~\cite{Graves12, GravesMohamedHinton13} as
a generalization of connectionist temporal classification
(CTC)~\cite{GravesFernandezGomezEtAl06}.
The RNN-T model defines a probability distribution over the output label
sequences conditioned on the input acoustics, $P(\y|\x)$, by marginalizing over
all possible alignment sequences, $\yhat$.
Specifically, RNN-T introduces a special blank symbol, $\blank$, and defines the
set of all valid frame-level alignments, $\mathcal{B}(\x, \y)$, as the set of
all label sequences, $\yhat = (\hat{y}_1, \ldots, \hat{y}_{T+U})$, where
$\hat{y}_i \in \mathcal{Y} \cup \left\{\blank\right\}$, such that $\yhat$ is
identical to $\y$ after removing all $\blank$ symbols.

The RNN-T model is depicted in Figure~\ref{fig:rnnt}.
As can be seen in the figure, the model consists of three components: an
\emph{acoustic encoder} (a stack of unidirectional LSTM
layers~\cite{HochreiterSchmidhuber97}, in this work) which transform the input
acoustic sequence into a higher-level representation, $\henc = [\henc_1, \ldots,
\henc_T]$; a \emph{prediction network} (another stack of unidirectional LSTMs,
in this work); and a \emph{joint network} which combines these to produce a
distribution over the output symbols and $\blank$:
\begin{align}
  P(\y | \x) &= \sum_{\yhat \in \mathcal{B}(\x, \y)} P(\yhat|\x)
	% \\
	% &= \sum_{\yhat \in B(\x, \y)} \prod_{i=1}^{T+U} P(\hat{y}_i | \x_1, \ldots, \x_{t_i}, y_0, \ldots, y_{u_i}) \label{eqn:rnnt}
  &= \sum_{\yhat \in \mathcal{B}(\x, \y)} \prod_{i=1}^{T+U} P(\hat{y}_i | \x^{t_i}_1, y^{u_i}_0) \label{eqn:rnnt}
\end{align}
\noindent where, $\x^{t_i}_1 = [\x_1, \ldots, \x_{t_i}]$, $y^{u_i}_0 = [y_0,
\ldots, y_{u_i}]$, and $y_0 = \sos$ is a special symbol denoting the start of the sentence;
$u_i$ and $t_i - 1$ denote the number of non-blank and blank symbols
respectively in the partial alignment sequence $\hat{y}_1, \ldots,
\hat{y}_{i-1}$.
Note that the prediction network is only input with non-blank symbols.
The summation in Equation~\ref{eqn:rnnt} and the gradients of the log-likelihood
function can be computed using the forward-backward algorithm~\cite{Graves12}.

\section{Limiting Prediction Network Context}
\label{sec:limited_context}
A vanilla RNN-T model is conditioned on all previous predictions, and can thus
be thought of as a \emph{full-context} model.
We can limit the context of the RNN-T model by modifying the prediction network
to only depend on a fixed number of previous labels.
Specifically, a model with $K$-gram context is conditioned on at most $K-1$
previous labels so that the output distribution of the RNN-T computes:
$P(\hat{y}_i | \x^{t_i}_1, y^{u_i}_{u_i - K +2})$.
Since our baseline full-context RNN-T model uses a stack of LSTM layers to model
the prediction network, in this work we limit the context through the use of an
LSTM-based prediction network that is reset and sequentially fed only the
sequence of the last $K-1$ labels at each step.
This ensures that our results are comparable to the baseline configuration.
However, other choices would also be reasonable to model a limited context
prediction network: e.g., a transformer as in~\cite{ZhangLuSakEtAl20}, or a
simple feed-forward network with $K-1$ inputs.
Each of these choices involves a different tradeoff in terms of computation
versus runtime memory usage, and we leave the study of these alternate
architectures for future work.

\subsection{Decoding with Path Merging to Create Lattices}
\label{sec:path_merging}
\begin{figure*}
  \centering
	\begin{subfigure}[t]{0.49\textwidth}
    \includegraphics[width=\textwidth]{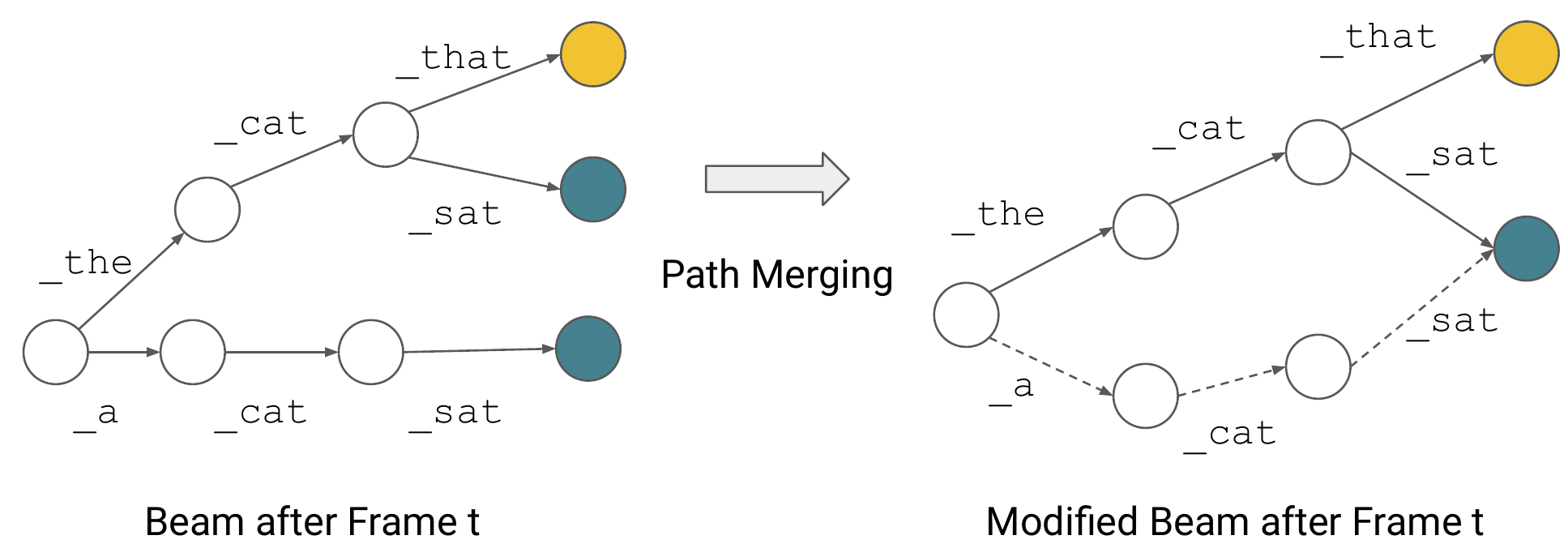}
    \caption{Path merging in a 3-gram limited context model.}
    \label{fig:limitedcontext_pathmerge}
    \end{subfigure}
    ~~
    \begin{subfigure}[t]{0.49\textwidth}
    \includegraphics[width=\textwidth]{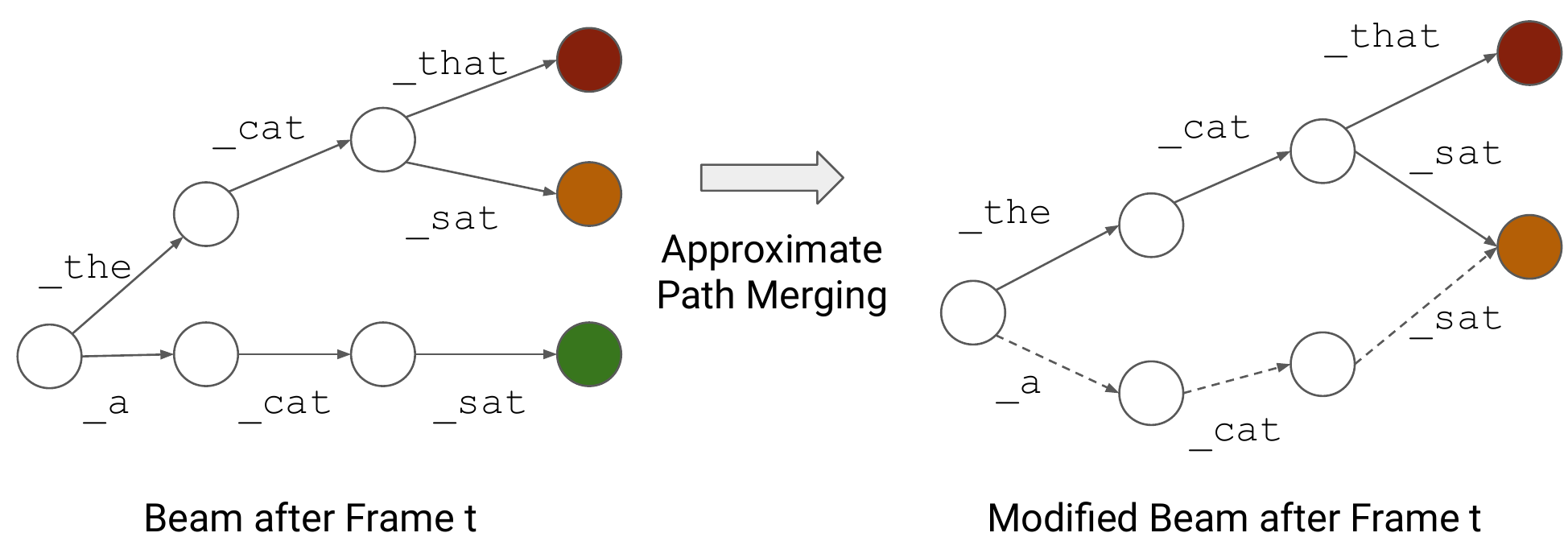}
    \caption{Approximate path merging in the full-context RNN-T model. The lower
	    cost path has been retained on the active beam.}
    \label{fig:baseline_pathmerge}
  \end{subfigure}
  \caption{Path-merging to create a lattice during decoding for a 3-gram model
	(left) and a full-context model (right). Colors represent the model
	state. Dashed lines represent paths that are included in the lattice,
	but removed from the active beam.}
  \label{fig:path_merging}
\end{figure*}
Traditional decoding algorithms for RNN-T~\cite{Graves12, TripathiLuSakEtAl19}
only produce trees that are rooted at the $\sos$ label since distinct
label sequences result in unique model states (i.e., the state of the prediction
network, since the encoder state is not conditioned on the label sequence).
In a limited context model, however, model states are identical if two paths on
the beam share the same local label history.
This allows for additional optimizations during the search, as illustrated in
Figure~\ref{fig:limitedcontext_pathmerge} for a 3-gram limited context model.
In this case, the model states for the partial hypotheses \emph{`a cat sat'} and
\emph{`the cat sat'} are identical (indicated by using the same color to
represent both states).
For this example, assume that \emph{`the cat sat'} is a lower cost (i.e., higher
probability) partial path.  Since future costs of identical labels sequences
starting from the blue states are identical, the current lower cost hypothesis
is guaranteed to be better than the current higher cost hypothesis for all
future steps.
Therefore, we can remove the higher cost path from the active beam, and instead
merge it with the lower cost path to create a lattice.
This has the effect of `freeing up' space on the beam, while retaining the
alternative paths in the final lattice where they can be used for downstream
applications.
Note that this can have a large impact since end-to-end models are typically
decoded with small number of candidates in the beam for
efficiency~\cite{SainathHeLiEtAl20}, and thus the beam diversity tends to reduce
for longer utterances~\cite{NarayananPrabhavalkarChiuEtAl19}.
We note that a similar mechanism has been proposed previously by Zapotoczny
\emph{et al.}~\cite{ZapotocznyPietrzakLancuckiEtAl19} in the context of lattice
generation for attention-based encoder-decoder models, and by Liu \emph{et
al.}~\cite{LiuWangChenEtAl14} in the context of efficiently rescoring lattices
with neural LMs.
To the best of our knowledge, our work is the first to apply these ideas to
streaming end-to-end models such as RNN-T to optimize the search process.

In contrast, in a full-context RNN-T model, illustrated in
Figure~\ref{fig:baseline_pathmerge}, model states are distinct for two partial
hypotheses if they correspond to distinct label sequences (represented using
distinct colors in the figure).
For full-context models, at least in principle, a higher cost partial path at an
intermediate point in the search could eventually be part of the lowest cost
complete path.
However, it is still possible to employ the same path merging scheme in
full-context RNN-T models through an \emph{approximation} if two partial
hypotheses share the \emph{same local history}, by retaining the model state
corresponding to the lower cost partial hypothesis as illustrated in the figure.
Note that in this case the retained state still corresponds to the full context
of the entire label sequence until that point.
As we demonstrate in Section~\ref{sec:results}, our proposed path-merging scheme
results in a more efficient search process while improving word error rate.

%% file: experiments.tex
\section{Experimental Setup}
\label{sec:experiments}
\textbf{Model Architecture:} Our experimental setup is similar to our previous
work~\cite{SainathHeLiEtAl20}.
The input acoustics are parameterized using 128-dimensional log-mel filterbank
energies computed over the 16KHz range
following~\cite{NarayananPrabhavalkarChiuEtAl19}, with 32ms windows and a 10ms
hop.
In order to reduce the effective frame rate, features from four adjacent frames
are concatenated together (to produce 512 dimensional features), which are
further sub-sampled by a factor of 3, so that the effective input frame rate is
30ms.
In this work, we also apply SpecAugment masks~\cite{ParkChanZhangEtAl19} using
the configuration described in~\cite{ParkZhangChiuEtAl20}, which we find to
improve performance over the system in~\cite{SainathHeLiEtAl20}.
The encoder network in all of our experiments is modeled using
a stack of 8 unidirectional LSTM~\cite{HochreiterSchmidhuber97} layers, each of
which contains 2,048 units and a projection layer of 640 units.
We add a time-reduction layer after the second LSTM layer which stacks two
adjacent outputs and sub-samples them by a factor of 2, so that the effective
encoder frame rate is 60ms.
The prediction network is modeled using two layers
of unidirectional LSTMs~\cite{HochreiterSchmidhuber97}, each of which contains
2,048 units with a projection layer of 640 units.
The joint layer is modeled as a single-layer feed-forward network
with 640 units.
The output of the joint network produces a distribution over 4,096
word-pieces~\cite{SchusterNakajima12} which are derived from a large text
corpus.
In total, each of our models contains $\sim$120 million trainable
parameters.

\textbf{Training Sets:} Our models are trained on a diverse set of utterances
from multiple domains including voice search, telephony, far-field, and YouTube
~\cite{NarayananPrabhavalkarChiuEtAl19}.
All utterances are anonymized and hand-transcribed; YouTube transcriptions are
obtained using a semi-supervised approach~\cite{LiaoMcDermottSenior13}.
In order to improve robustness to environmental distortions, models are trained
with additional noisy data using a room simulator~\cite{KimMisraChinEtAl17}.
Noise samples are drawn from YouTube and from daily life noisy recordings.
The noisy data are generated with an SNR of between 0dB and 30dB, with an
average SNR of 12dB; T60 reverberation times range from 0 -- 900ms with an
average of 500ms.
We employ mixed-bandwidth training~\cite{LiYuHuangEtAl12}, by randomly
downsampling the training data to 8KHz 50\% of the time.

\textbf{Test Sets:} Results are reported in two test domains: the first consists of
utterances drawn from Google voice search traffic (VS: 11,585 utterances; 56,069
words); the second set consists of data drawn from non-voice search Google
traffic (NVS: 12,426 utterances; 127,105 words).
The utterances in the NVS set tend to be longer (both in terms of duration and
label token length) on average than utterances in the VS set.
The 90th percentile token sequence length is $\sim$14 word-pieces for the VS set, and
$\sim$28 word-pieces for the NVS set.
Results are also reported \emph{more challenging versions} of the VS (VS-hard:
9,662 utterances; 46,673 words) and NVS test sets (NVS-hard: 19,411 utterances;
198,215 words) that contain more variation in terms of background noise, volume,
and accents.
All test set utterances are anonymized and hand-transcribed.

\textbf{Training:} Models are trained using Lingvo~\cite{ShenNguyenWuEtAl19} in
Tensorflow~\cite{AbadiBarhamChenEtAl16}, with $8\times8$ Tensor Processing Units
(TPUs)~\cite{JouppiYoungCliffEtAl17}.
Models are optimized using synchronized stochastic gradient descent with
mini-batches of 4,096 utterances using the Adam optimizer~\cite{KingmaBa14}.

\textbf{Full and Limited Context Models:}
In addition to the full-context RNN-T model (baseline), we study the impact of
limiting context by training models with varying amounts of context ranging from
2--10 (lc-2gram -- lc-10gram; i.e., conditioning on 1 -- 9 previously predicted
labels, respectively).

\textbf{Decoding:} All results are reported after decoding models using the
breadth-first search decoding
algorithm~\cite{TripathiLuSakEtAl19}.\footnote{Similar results are obtained using
the best-first search decoding algorithm~\cite{Graves12}, however these are
omitted due to space limitations.}
The limited context models are always evaluated using the path-merging process
proposed in Section~\ref{sec:path_merging}.
We also consider the approximate path-merging scheme described in
Section~\ref{sec:path_merging} applied to the baseline during inference.  Note
that in this case, the baseline is always \emph{trained with full-context};
during evaluation, we merge paths if the last 1 -- 9 non-blank labels are
identical for two paths (base-pm2 -- base-pm10), and retain the state
corresponding to the lower cost path.
All models are decoded with a maximum beam size of 10 partial hypotheses and a
local beam of 10 (i.e., maximum allowable absolute difference between
log-likelihoods of partial hypotheses).

\textbf{Evaluation Metrics:} Results are reported in terms of both word error
rate (WER) as well as the oracle WER in the lattice (for the systems with
path-merging) or the N-best list (for the baseline configuration).
In order to evaluate whether the proposed path merging scheme described in
Section~\ref{sec:path_merging} improves computational efficiency, results are
also reported in terms of the number of model states which are expanded during
the search.
Specifically, we report the average number of joint network evaluations per
utterance for each of the test sets (average model states), which serve as a
good proxy to measure the cost of the search.

%% file: results.tex
\section{Results}
\label{sec:results}
\begin{table}[t]
  \centering
  \begin{tabular}{|c|c|c|c|c|}
    \hline
    \multirow{2}{*}{\textbf{System}} & \multicolumn{4}{|c|}{\textbf{WER / Oracle
	  WER (\%)}} \\
    \cline{2-5}
	  & {VS} & {NVS} & {VS-hard} & {NVS-hard} \\
    \hline
    baseline & \textbf{6.0} / 1.7 & 3.3 / 1.1 & 7.6 / 2.8 & 7.4 / 3.9 \\
    \hline
    lc-2gram & 6.4 / \textbf{1.1} & 3.5 / 0.7 & 8.1 / 2.2 & 8.0 / 3.3 \\
    lc-3gram & 6.2 / 1.2 & 3.3 / 0.7 & 7.8 / 2.2 & 7.6 / 3.1 \\
    lc-4gram & 6.1 / 1.3 & 3.3 / 0.7 & 7.6 / 2.2 & \textbf{7.2} / 3.0 \\
    lc-5gram & \textbf{6.0} / 1.3 & \textbf{3.2} / 0.7 & 7.6 / 2.4 &
    \textbf{7.2} / \textbf{2.9}\\
    lc-7gram & \textbf{6.0} / 1.5 & \textbf{3.2} / 0.8 & 7.7 / 2.6 & 7.3 / 3.2 \\
    lc-10gram & \textbf{6.0} / 1.7 & 3.3 / 0.9 & \textbf{7.5} / 2.7 & \textbf{7.2} / 3.4 \\
    \hline
    base-pm2 & 6.2 / \textbf{1.1} & 3.4 / \textbf{0.6} & 7.9 / \textbf{2.1} & 7.6 / 3.1 \\
    base-pm3 & 6.1 / \textbf{1.1} & 3.3 / \textbf{0.6} & 7.7 / \textbf{2.1} & 7.5 / 3.1 \\
    base-pm4 & 6.1 / 1.3 & 3.3 / \textbf{0.6}  & 7.7 / 2.3 & 7.4 / 3.0 \\
    base-pm5 & \textbf{6.0} / 1.4 & \textbf{3.2} / 0.7 & 7.6 / 2.4 & 7.4 / 3.1 \\
    base-pm7 & \textbf{6.0} / 1.5 & \textbf{3.2} / 0.8 & 7.7 / 2.5 & 7.3 / 3.3 \\
    base-pm10 & \textbf{6.0} / 1.6 & 3.3 / 0.9 & \textbf{7.5} / 2.7 &  \textbf{7.2} / 3.4 \\
    \hline
  \end{tabular}
  \caption{WERs and Oracle WERs (\%) for the systems described in
	Section~\ref{sec:experiments}. The systems labeled lc-Ngram are trained
	with limited context and evaluated with path merging; systems labeled
	base-pmN correspond to the baseline which is trained with full-context,
	but evaluated with path-merging. The best performing systems are
	indicated in \textbf{bold font}.}
  \label{tbl:results}
\end{table}
Our results are presented in Table~\ref{tbl:results}.
First, we consider the systems \emph{trained} by limiting prediction network
context and evaluated with path-merging.
As can be seen in the table, a model with 5-gram context (i.e., conditioned on
four previous labels) performs as well as the baseline model across all test
sets.
In fact, this model outperforms the baseline on the NVS and NVS-hard sets which
tend to contain longer utterances.
We hypothesize that this is likely a consequence of our proposed path merging
strategy which removes redundant paths from the beam, thus allowing the model to
represent diverse competing hypotheses as described in
Figure~\ref{fig:path_merging} and Section~\ref{sec:path_merging}.
This is further supported by comparing the Oracle WERs of the limited context
systems relative to the baseline.
Models with smaller contexts have more opportunities for path-merging, and thus
achieve lower oracle WERs relative to models with larger contexts; this comes,
however, at the cost of a degradation in the WER.
Using 5-gram contexts provides the best balance between the two evaluation
metrics, allowing the model to improve oracle WERs by between 14.3--36.4\%
across the various test sets, with larger improvements on the longer
NVS/NVS-hard sets.
We also note that our results are in contrast to previous studies with have been
conducted on smaller-scale tasks with limited training
data~\cite{ZhangLuSakEtAl20, GhodsiLiuApfelEtAl20}.

Second, we consider the baseline model which is trained with full-context, but
evaluated with path-merging using varying amounts of context.
From the results in Table~\ref{tbl:results}, we observe that the baseline
models evaluated with path-merging perform at least as well the full-context
baseline if not better, as long as we use at least 5-gram context during path
merging.
As was the case with the models trained with limited-context, using path-merging
with 5-gram contexts appears to provide the best WER relative to the baseline
configuration.
Finally, we note that these systems achieve similarly large Oracle WER
improvements as the systems trained with limited context.

\subsection{Reducing Search Complexity Through Path-Merging}
\begin{table}[t]
  \centering
  \begin{tabular}{|c|c|c|c|c|c|}
    \hline
    \multirow{2}{*}{\textbf{System}} & \multicolumn{5}{|c|}{\textbf{Average Model States}} \\
    \cline{2-6}
	  & {VS} & {NVS} & {VS-hard} & {NVS-hard} & {Avg} \\
    \hline
	  baseline   & 770.3 & 1824.9 & 766.3 & 2005.6 & 1342.9 \\ % 972.6 & 2104.4 & 978.9 & 2362.8 & 1604.7 \\
    \hline
	  lc-2gram   & 652.7 & 1344.1 & 662.7 & 1601.7 & 1065.3 \\ % 846.6 & 1579.2 & 872.2 & 1925.7 & 1306.4 \\
	  lc-3gram   & 694.5 & 1560.7 & 697.2 & 1791.8 & 1186.0 \\ % 945.7 & 2104.4 & 953.4 & 2351.9 & 1588.4 \\
	  lc-4gram   & 725.3 & 1668.3 & 728.6 & 1857.9 & 1245.1 \\ % 955.2 & 2105.4 & 967.5 & 2337.5 & 1591.4 \\
	  lc-5gram   & 732.9 & 1711.9 & 735.9 & 1906.8 & 1271.9 \\ % 926.9 & 1973.7 & 943.3 & 2251.3 & 1523.8 \\
	  lc-7gram   & 759.0 & 1764.8 & 762.7 & 1935.0 & 1305.4 \\ % 961.9 & 2033.9 & 978.9 & 2280.7 & 1563.9 \\
	  lc-10gram  & 771.2 & 1816.9 & 763.0 & 1977.5 & 1332.1 \\ % 977.3 & 2095.3 & 977.6 & 2331.4 & 1595.4 \\
    \hline
	  base-pm2  &  643.6 & 1323.3 & 651.9 & 1587.5 & 1051.6 \\ % 823.7 & 1546.0 & 844.1 & 1897.2 & 1277.7 \\
	  base-pm3  &  712.3 & 1549.8 & 714.1 & 1784.5 & 1190.2 \\ % 904.9 & 1794.5 & 917.9 & 2116.0 & 1433.3 \\
	  base-pm4  &  739.5 & 1653.9 & 738.4 & 1869.4 & 1250.3 \\ % 937.3 & 1909.2 & 947.0 & 2210.6 & 1501.1 \\
	  base-pm5  &  753.0 & 1712.1 & 750.4 & 1914.9 & 1282.6 \\ % 953.2 & 1973.3 & 961.1 & 2261.4 & 1537.2 \\
	  base-pm7  &  763.4 & 1768.9 & 760.0 & 1957.9 & 1312.6 \\ % 965.1 & 2036.3 & 972.1 & 2309.6 & 1570.8 \\
	  base-pm10 &  768.2 & 1801.8 & 764.2 & 1982.9 & 1329.3 \\ % 970.4 & 2073.0 & 976.7 & 2337.4 & 1589.4 \\
    \hline
  \end{tabular}
	\caption{Average number of model states (i.e., the average number of
	joint network evaluations) per utterance for the systems described in
	Section~\ref{sec:experiments}. The \textbf{Avg} column corresponds
	to the average over the four test sets. Lower values indicate a more
	efficient search.}
  \label{tbl:model_states}
\end{table}
In Table~\ref{tbl:model_states} we measure the complexity of the search process
for each of the systems in Table~\ref{tbl:results}.
For the purposes of this analysis, all models are evaluated with exactly the
same pruning parameters as described in Section~\ref{sec:experiments}.
As can be observed in the table, the use of path merging consistently results in
a more efficient search -- this is true for both systems trained with limited
context (i.e., lc-2gram -- lc-10gram) as well as for the baseline (i.e.,
base-pm2 -- base-pm10).
In general, systems which merge hypotheses more aggressively (e.g., lc-2gram)
result in fewer average model states then systems which merge hypotheses less
aggressively (e.g., lc-10gram).
Comparing the complexity of the search for the baseline relative to the two best
performing configurations in terms of WER -- namely lc-5gram and base-pm5 -- we
observe that path merging improves efficiency by 5.3\% and 4.5\%, respectively
without any degradation in WER.

%% file: conclusions.tex
\section{Conclusions and Discussion}
\label{sec:conclusions}
In this work, we studied the impact of label conditioning in the RNN-T model on
WER and its implications for improving the efficiency of the search.
In experimental evaluations, we found that a full-context RNN-T model with
word-piece outputs performs comparably to a model trained with context limited
to four previous labels.
We also investigated modifications to the decoding strategy which are enabled by
limiting context: either exactly -- by training a model with limited context; or
approximately in the baseline -- by merging paths during decoding to produce
lattices if two paths share the same local label history.
The proposed path-merging strategy was shown to improve the oracle WER in the
lattice by up to 36\%, while improving the efficiency of the search by reducing
the number of model evaluations by up to 5\% without any degradation in WER.

The ability to create dense lattices which represent alternative hypotheses has
a number of potential applications; future work will investigate its
effectiveness in improving word-level confidence estimates~\cite{KempSchaaf97},
and in improving WER by rescoring lattices in the
second-pass~\cite{SainathPangRybachEtAl19} taking advantage of the improved
oracle error rates.
Finally, we note that our work opens up interesting research directions for new
RNN-T architectures which limit prediction network context to just a few
previous labels; replacing recurrent LSTM units through such a process can
reduce sequential dependencies in the system, thus potentially improving
execution speed.